\documentclass{goodfire}
\usepackage{goodfire}

\usepackage{noto-sans}

\renewcommand{\sfdefault}{\cabinfamily} 

\usepackage{amsmath,amsfonts,bm}

\def\eqref#1{equation~\ref{#1}}

\def\1{\bm{1}}

\def\vc{{\bm{c}}}

\def\vu{{\bm{u}}}

\def\vx{{\bm{x}}}

\DeclareMathAlphabet{\mathsfit}{\encodingdefault}{\sfdefault}{m}{sl}
\SetMathAlphabet{\mathsfit}{bold}{\encodingdefault}{\sfdefault}{bx}{n}

\usepackage{graphicx} 
\usepackage{float}
\usepackage[export]{adjustbox}
\usepackage{amsmath, amssymb}
\usepackage{natbib}

\usepackage{microtype}
\usepackage{siunitx}
\usepackage{booktabs}
\usepackage{hyperref}
\usepackage{xurl}
\usepackage{cleveref}
\usepackage{wrapfig}
\usepackage{enumitem}
\usepackage[color=pink!40, textsize=tiny]{todonotes}

\usepackage{pgfplots}
\usepgfplotslibrary{groupplots,colormaps}
\usetikzlibrary{hobby,arrows.meta}

\tikzset{font=\footnotesize}
\pgfplotsset{compat=newest, width=0.9\textwidth, height=4cm}

\title{Language Model Activations Inhabit Privileged Error-Correcting Basins}

\usepackage[subdued,defaultmathsizes,italic,LGRgreek+]{mathastext}

\MTgreekfont{\notosansfamily}
\MTDeclareVersion{figurecaption}{T1}{\cabinfamily}{m}{n}
\DeclareCaptionFont{captionmath}{%
  \MTversion*{figurecaption}\let\bm\mathnormalbold
  \sisetup{mode=text,text-font-command=\cabin}%
}
\AddToHook{env/tabular/begin}{\sffamily\sisetup{mode=text,text-font-command=\notosans}}
\AddToHook{env/tabular*/begin}{\sffamily\sisetup{mode=text,text-font-command=\notosans}}

\graphicspath{{fig/goodfire/}{fig/shared/}}

\abstract{Language models exhibit remarkable robustness, continuing to produce coherent text even when their activations are perturbed by interventions like linear steering.
We hypothesize that this robustness is a result of passive dynamics,
i.e., constraining mechanisms in the forward pass that funnel activations toward ``good'' regions that produce coherent outputs.
To investigate these hypothesized error-correcting mechanisms,
we probe the geometry of language model activation space
by observing the action of model layers on low-dimensional curves.
In doing so, we discover that model activations occur within a cluster of distinct attracting basins,
which differentiate natural activations geometrically from distributionally similar synthetic activations.
Applying this lens to language model steering,
we observe feature-specific basins along semantic steering directions,
and find that steering moves activations between these basins.
To demonstrate the active role of this geometry in neural computation,
we show that adaptively modulating steering strength to transport activations across basins improves inter-language steering,
significantly increasing the probability of sampling tokens from the target language
compared to fixed-strength steering.
Our findings establish analysis of activation space geometry as a promising approach to interpreting and controlling language models.
}
\authors{
  Matthew Finlayson,\textsuperscript{1,2}
  Francisco Pernice,\textsuperscript{1,3}
  Eric Todd,\textsuperscript{1,4}
  Amir Zur,\textsuperscript{1,5}
  Daniel Wurgaft,\textsuperscript{1,5}
  Fenil~R.~Doshi,\textsuperscript{1,6}
  Vasudev Shyam,\textsuperscript{1}
  Matt Feiszli,\textsuperscript{1}
  Satchel Grant,\textsuperscript{1}
  Lucius Bushnaq,\textsuperscript{1}
  Tal Haklay,\textsuperscript{1}
  Usha~Bhalla,\textsuperscript{1}
  Matthew~Kowal,\textsuperscript{1}
  Thomas Fel,\textsuperscript{1}
  Jack Merullo,\textsuperscript{1}
  Atticus Geiger,\textsuperscript{1}
  Xiang Ren,\textsuperscript{2}
  Owen Lewis,\textsuperscript{1}
  \&~Ekdeep~Singh~Lubana\textsuperscript{1}
  }
\affiliations{
  \textsuperscript{1}Goodfire \quad
  \textsuperscript{2}University of Southern California \quad
  \textsuperscript{3}Massachusetts Institute of Technology \quad
  \textsuperscript{4}Northeastern University \quad
  \textsuperscript{5}Stanford University \quad
  \textsuperscript{6}Harvard University
}

\begin{document}

\maketitlebox


\section{Introduction}

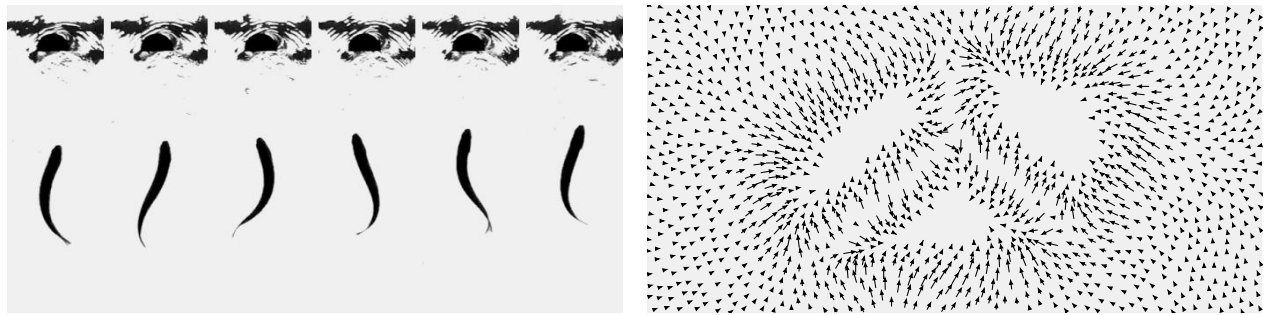
\begin{figure}[b]
  \centering
  \includegraphics[width=.49\linewidth,valign=T]{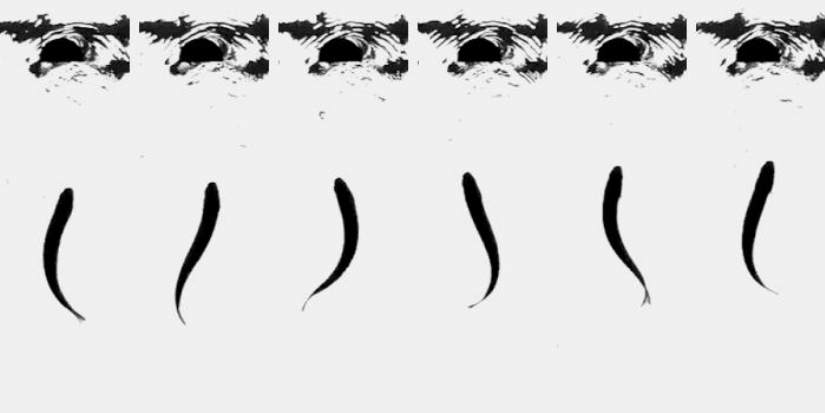}%
  \hfill
  \begin{tikzpicture}[baseline=(current bounding box.north)]
    \definecolor{axisbackground}{RGB}{240,240,240}
    \begin{axis}[
        width=.49\linewidth,
        height=.245\linewidth,
        scale only axis,
        axis equal image,
        xmin=-2.8, xmax=2.8,
        ymin=-1.4, ymax=1.4,
        enlargelimits=false,
        axis background/.style={fill=axisbackground},
        colormap/viridis high res,
        point meta min=0, point meta max=0.75,
        hide axis,
    ]
      \addplot[
        point meta=\thisrow{speed},
        black,
        quiver={
          u={\thisrow{u}},
          v={\thisrow{v}},
          scale arrows=0.177,
        },
        -{Latex[scale=0.5]},
      ] table[x=x, y=y] {data/two_basin_field.dat};
      \path (rel axis cs:0,0) rectangle (rel axis cs:1,1);
    \end{axis}
  \end{tikzpicture}
  \caption{
    Biological organisms use passive dynamics to reduce the cost of explicit bodily control; for instance,
    a trout corpse swims upstream in the presence of turbulence~\citep[left,][]{BEAL_HOVER_TRIANTAFYLLOU_LIAO_LAUDER_2006}.
    We speculate that the passive dynamics within language models' latent space
    similarly constrain the model toward coherent outputs (right schematic).
    We advance this theory through a set of experiments
    that characterize the latent space geometry as
    a set of basins that attract activations.
    We demonstrate the role of these basins in neural computation
    by showing that linear steering transports activations between basins.
  }
   \label{fig:coast}
\end{figure}

In biology, it is often advantageous for an organism's physical morphology to minimize the amount of control required to produce its movements.
For instance, \citet{126245} shows that knees allow a pair of humanoid legs to passively walk down a slope without any muscular input,
and \citet{BEAL_HOVER_TRIANTAFYLLOU_LIAO_LAUDER_2006} document how the biomechanical constraints of a trout corpse cause it to passively swim upstream in the presence of turbulence.
Similarly, the constraints imposed by a marionette's joints allow a puppeteer to produce lifelike movements with minimal control, i.e., using only strings.
This phenomenon, known as morphological computation, allows organisms to offload the cost of explicit control to their physiology and interactions with the environment~\citep{PAUL2006619}.

We propose that an analogous phenomenon is at play within large language models:
the model's layers constrain the latent space such that the model is virtually guaranteed to produce a coherent continuation under almost any input.
We hypothesize that this is accomplished via 
an error-correcting dynamic,
wherein the model layers distort the latent space to funnel activations toward a set of attracting basins \citep{hopfield1982neural, khona2022attractor, sussillo2013opening, hänni2024mathematicalmodelscomputationsuperposition}.
Such a mechanism would allow the model to recover gracefully from coarse manipulations like linear steering \citep{turner2023steering, panickssery2024steeringllama2contrastive}.

To support this hypothesis, we perform experiments that identify activation basins within the model
and characterize their geometry.
Prior work has found evidence of error correction \citep{ rushing2024explorations, mcgrath2023hydra, patrawala2025llmlayersimmediatelycorrecteachother}
and stable regions in the residual stream~\citep{janiak2024characterizing, da2026evidence, shinkle2025activation, giglemiani2024evaluatingsyntheticactivationscomposed}. We build upon these works, further characterizing the space as a cluster of distinct, closely-packed attracting basins
which appear to constitute a feasible region,
i.e., all activations from natural text appear to occur within the basin cluster. 
Naturally occurring activations 
are geometrically distinguishable from distributionally similar synthetic ones;
in particular, synthetic activations within the basin cluster undergo norm shrinkage as they pass forward through the model, while natural activations exhibit moderate norm growth.\footnote{Concretely, for model block $f$ and activation $\vx$, the norm shrinking or growing means $\lVert f(\vx)\rVert_2$ is smaller or larger than $\lVert \vx\rVert_2$.}
As synthetic activations pass through additional layers, they gradually revert
to moderate norm growth, which we interpret as further evidence of a distributed error correcting mechanism.

Our characterization of the latent space suggests that linear steering,
i.e., adding a fixed vector to activations to modify model behavior \citep{turner2023steering, panickssery2024steeringllama2contrastive},
may be understood as transporting activations from one basin to another.
As an illustrative case study, we show that \model{Llama 3 8B Instruct}
activation space has a pair of basins
along the feature direction corresponding to refusal and compliance, 
and that the activations from refusal responses \citep{arditi2024refusal} fall into one basin, 
while activations from compliant responses fall into the other.
Steering an activation into the refusal basin at a single token position is sufficient to trigger the model to refuse a benign request.
We also identify feature basins for different languages, using English, Chinese, and Greek, and we show that
the geometry of the feature basins hints at a hierarchical representation space that privileges English \citep{wendler2024llamas}.
In particular, the English basin occupies a much larger angular region, while other languages share the remainder of the activation~space.

To establish the role of activation basins in neural computation, 
we leverage basins to make linear model steering more effective. 
Steering often requires tuning the steering strength (i.e., scaling a fixed steering vector) to achieve the desired influence and avoid model degradation. 
It also does not take into account the activation space geometry.
Recent work has therefore explored adaptive activation steering and geometry-aware steering~\citep{wang2025adaptive, pmlr-v267-hedstrom25a, ferrando2025dynamically, ICLR2025_df4f6e43, raval2026curveballsteeringrightdirection, wurgaft2026manifold, oozeer2026riemannianmanifoldsteeringgeometryawaregenerative, wang2026fishbackpullbackfishergeometry}.
We show that adaptively modulating the steering strength on a per-instance basis,
so that the perturbed activation lands within the target feature basin,
significantly improves effectiveness over fixed-strength steering.
In the English-to-Chinese steering setting, our parameter-free method yields a 19\% reduction in non-Chinese and 4\% increase in Chinese tokens under greedy decoding.

Our concrete contributions include:
\begin{itemize}[nosep]
  \item Geometrically characterizing privileged attracting basins within the activation space.
  \item An account of model steering as transporting activations between feature basins.
  \item A parameter-free approach to select steering strength based on the activation space geometry.
\end{itemize}
These findings advance our understanding of language models' computation mechanisms from a macro-level, give a theory to explain model robustness, and demonstrate a path toward greater control over model behavior that leverages the geometry of the activation space.

\section{Probing activation space geometry}
\label{sec:method}

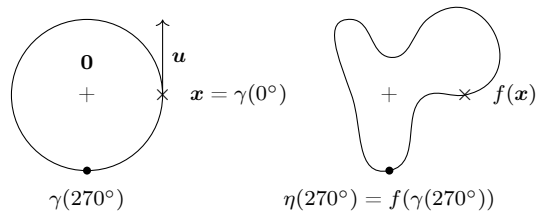
\begin{wrapfigure}{r}{0.5\textwidth}
\centering
\vspace{-20pt}
  \begin{tikzpicture}
    \draw (0,0) circle [radius=1];
    \node[label=$\bm{0}$] at (0,0) {+};
    \node[label=right:{$\vx=\gamma(\ang{0})$}] at (1,0) {$\times$};
    \draw[->] (1,0) -- node[right] {$\vu$} (1,1);
    \node[label=below:$\gamma(\ang{270})$] at (0,-1) (perturb) {};
    \fill (perturb) circle[radius=1.5pt];

    \begin{scope}[xshift=4.5cm]
      \draw (-1,0) to[closed, curve through={(-1,1) (-0.5,0.5) (0,1) (0.5,0) (0,0)}] (-0.5,-1);
      \node[label=right:$f(\vx)$] at (0.5,0) {$\times$};
      \node at (-0.5,0) {+};
      \node[label=below:{$\eta(\ang{270})=f(\gamma(\ang{270}))$}] at (-0.5,-1) (perturb) {};
      \fill (perturb) circle[radius=1.5pt];
    \end{scope}
  \end{tikzpicture}
\caption{\textbf{Toy Depiction of Spherical Probe.} We probe the activation space by observing the action of transformer blocks (\(f\)) on low-dimensional spheres through activations \(\vx\) centered at the origin. Applying the block to every point on the sphere yields a closed low-dimensional surface embedded in \(\mathbb{R}^d\).
Parametrizing the sphere by angle \(\theta\),
we can analyze the output curve and its derivatives 
to identify attracting basins and other features.
}
  \label{fig:sphereprobe}
  \vspace{-10pt}
\end{wrapfigure}

To probe language model activation space we make use of
circular paths centered at the origin.\footnote{
We use sphere-based probes to preserve the norm across the path.
Preliminary exploration with linear probes proved difficult due to large-norm activations at the extreme point of the paths asymptotically approaching $f(\vx)=\vx+\vc$ for some constant $\vc$ in models that normalize the residual update.}
As shown in \cref{fig:sphereprobe}, we generally choose the circle to pass through some activation of interest $\vx$
with tangent in unit direction $\vu$. 
We can parameterize this path as $\gamma(\theta)=\lVert\vx\rVert_2(\vu\sin\theta+\hat\vx\cos\theta)$.
Forwarding the path through a block $f$ of one or more model layers, 
we get a curve $\eta(\theta)=f(\gamma(\theta))$
which forms a closed loop embedded in \(\mathbb{R}^d\).
We analyze the resulting path in terms of 
$\eta(\theta)$ and its derivative $\dot\eta(\theta)$.
We sometimes generalize this 1-dimensional spherical probe to 2 dimensions by adding a second tangent direction at the activation.

The (potentially multivariate) derivatives of the parameterized surface~\(\eta\) constitute a measure of expansion.
Assuming that the derivative is sufficiently smooth and continuous,
if the value of \(\lVert\dot\eta(\theta)\rVert_2\) is large at some \(\theta\),
it means that points on the surface of the sphere become more spread out at the readout layer,
while small values mean the points become more densely packed.
In 2 dimensions, we can use the change in area of the infinitesimal tangent plane at a point on the sphere to capture the same notion.
We are primarily interested in the relative expansion
rates of different regions of the sphere.


From our own observations, and those of \citet{shinkle2025activation},
the error correction phenomenon we are interested in is primarily mediated by feedforward layers.
Therefore, for the sake of simplicity and to isolate the effect of the feedforward layers,
we ablate the context in select experiments, leaving only a beginning-of-sequence (\textsc{bos}) token.
This means that the attention layers can only attend to the previous and current token positions.\footnote{
We leave the \textsc{bos} token in context because some models, particularly Llama, use the first sequence token as an attention sink~\citep{xiao2023streamingllm}, so ablating the entire context results in the remaining token position behaving strangely, e.g., acquiring a massive norm~\citep{sun2024massive}.
}

To obtain natural activations, we randomly sample \(n\) token positions from random documents from \dataset{FineWeb} \citep{penedo2024the} and \dataset{The Pile}~\citep{gao2020pile800gbdatasetdiverse}.
For comparison, we also generate synthetic activations with varying levels of ``naturalness''.
We obtain \textit{norm-matched} synthetic activations by sampling random directions and randomly scaling by random Gaussian-distributed factors with the same mean and variance as the natural activation norms.
We obtain \textit{shuffled} activations by independently permuting each coordinate between the sampled natural activations,
retaining the mean and coordinate-wise variance of the natural activations.
We also obtain \textit{Gaussian} activations by taking the mean and covariance matrix of the random natural activations and sampling from a Gaussian with the same statistics.
We perform experiments on \model{Llama 3.1 8B}, \model{Llama 3 8B Instruct}~\citep{grattafiori2024llama3}, \model{OLMo 3 7B}~\citep{olmo2025olmo3}, and \model{Qwen 3 8B}~\citep{qwen3technicalreport}.

\section{Transformer block dynamics privilege natural activations}

In this section we explore the geometry of the latent activation space of language models using spherical probes as described in \cref{sec:method}.
We find evidence that natural language activations inhabit a privileged region within the activation space, meaning that the geometry induced by a block of one or more transformer layers deforms the space differently around a natural activation than it does for a synthetic activation.
We identify differences in terms of growth of the activation norms,
and find evidence of error correction dynamics whereby synthetic and perturbed activations move into natural activation regions over the course of the forward pass.

\subsection{Natural activations inhabit local norm growth maxima.}

\begin{figure}
  \centering
  \includegraphics{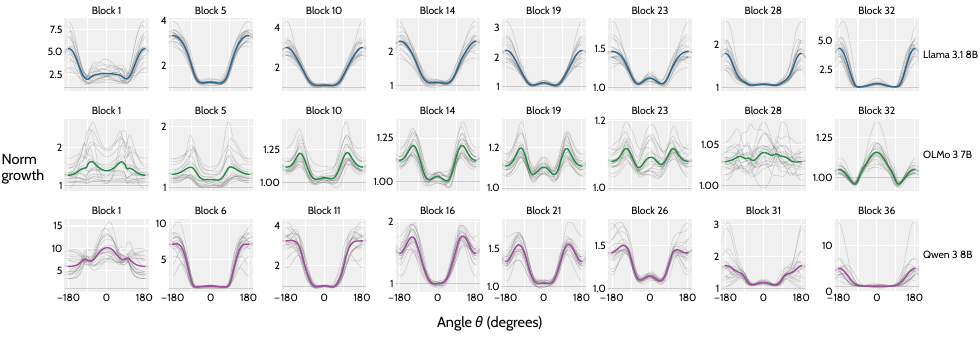}
  \caption{
    \textbf{Per-layer norm growth profiles for different models.} 
    Horizontal axis shows the perturbation angle \(\theta\) from the natural activation,
    wrapping around at \ang{180}.
    Norm growth is shown on vertical axis, 
    calculated $\lVert \eta(\theta)\rVert_2/\lVert\gamma(\theta)\rVert_2$. 
    Colored curves show average over 20 profiles, which are shown in gray.
    Tokens sampled from The Pile, context ablated. Non-ablated context results in \cref{fig:great_circle_norm_growth_context} are similar.
    All models show a distinctive norm-growth basin around the natural activations.
    The activations themselves often lie at a minor local norm growth maximum.
}
  \label{fig:great_circle_norm_growth}
\end{figure}

We find that language model activation space has a distinctive pattern of norm growth in the region around natural activations.
Using random tangent directions \(\vu\), 
we plot in \cref{fig:great_circle_norm_growth} the relative norm growth of points along the circle, i.e.,
\(\lVert\eta(\theta)\rVert_2/\lVert\gamma(\theta)\rVert_2\).
The region around the activations is characterized by modest norm growth, surrounded by high norm growth regions.
This is consistent across models and intermediate layers, except for the early and late layers for some models.
For several layers, the natural activation lies at the peak of a small local maximum within the basin.
For OLMo and some layers of Qwen, the activation's antipode~$-\vx$ at \ang{\pm180} also lies in a smaller radial basin.

\begin{figure}
  \centering
  \includegraphics{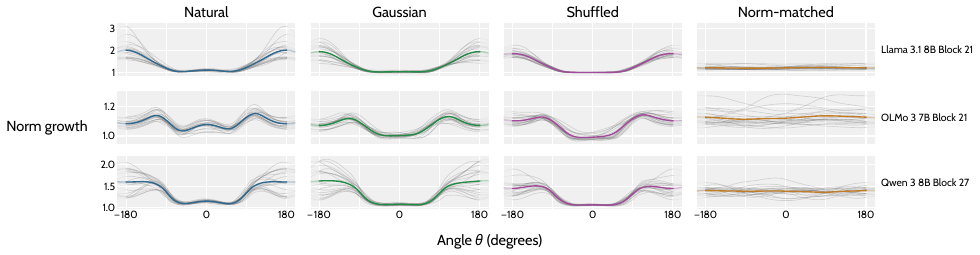}
  \caption{
    \textbf{Comparing natural to synthetic activations' norm growth profiles for various models.}
    Layers selected for high separation between Natural and Gaussian.
    Norm-matched synthetic activations do not fall within the norm-growth basin.
    Gaussian and shuffled activations fall within the norm growth basin, 
    but do not exhibit the local norm-growth maximum at \ang{0}, or, for Qwen, only exhibit a very small one.
  }
  \label{fig:great_circle_synthetic_growth}
\end{figure}

This radial norm growth profile is unique to natural activations,
meaning that it can distinguish natural activations from synthetic activations.
We compare the profiles between natural and norm-matched, Gaussian, and shuffled activations in \cref{fig:great_circle_synthetic_growth}.
Spheres through norm-matched random activations \textit{a priori}
have no privileged direction, since every direction is equally likely to be sampled,
making it easy to distinguish them from natural activations by checking whether the activation lies within a norm-growth basin.
Shuffled and Gaussian activations do show the same characteristic 
low-growth basin that natural activations exhibit,
but they do \emph{not} occur on local maxima within the growth basin.
A simple growth threshold separates them well at many layers, with maximum \textsc{auroc} for Gaussian activations (the harder to distinguish of the two) reaching 0.965 (Qwen)/0.983 (Llama)/0.997 (OLMo); see \cref{tab:fstorderclassifier} for full results.
In fact, synthetic activations tend to \emph{shrink},
i.e., have norm growth less than 1, at many layers.
This is remarkable, since it means that the residual update must have anti-alignment with the activation, which is unlikely to occur by chance in high dimensional space.
In contrast, natural activation norms generally grow uniformly as they pass through the model layers~\citep{heimersheim2023residual}.

\begin{figure}
  \centering
  \includegraphics{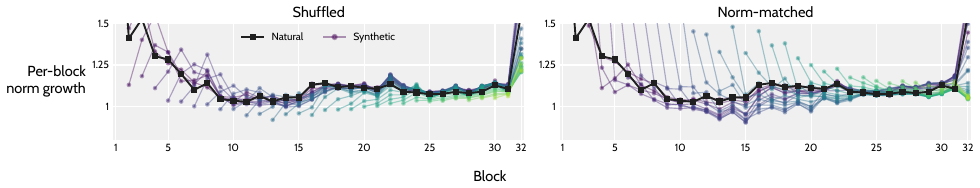}
  \caption{
    \textbf{Norm growth of natural and synthetic activations} as they pass forward through Llama.
    Each line represents the mean of a set of 128 sampled activations,
    color coded by the layer at which they are injected. 
    Natural activations norm growth shown in black.
    Synthetic activations gradually approach natural activation norm growth rates,
    suggesting that the model routes synthetic activations
    to more natural regions of the activation space.
    OLMo and Qwen in \cref{fig:traject}.
  }
  \label{fig:norm_growth_trajectory}
\end{figure}

\begin{figure}
  \centering
  \includegraphics{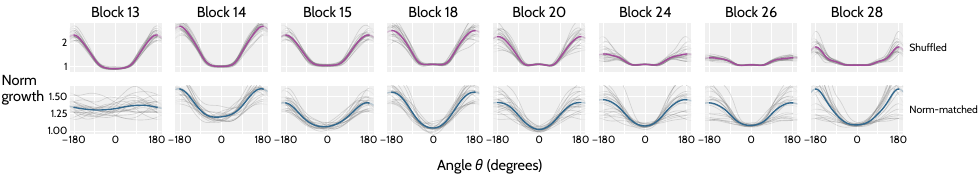}
  \caption{
    \textbf{Norm growth profile of synthetic activations as they pass through Llama layers 13--28.} Probes re-initialized at each layer. The norm-matched activation moves almost immediately into a norm growth basin, and the shuffled activation eventually moves to a local norm growth maximum, matching the profile of natural activations.
  }
  \label{fig:errorcorrect}
\end{figure}

We see evidence of error correction in the synthetic activations' norm growth as they pass forward through model layers.
In particular, synthetic norm growth approaches natural norm growth as they pass through more layers, as shown in \cref{fig:norm_growth_trajectory}.
This suggests that the activation has migrated to a more ``natural'' region of the basin cluster and is being treated accordingly by the model.
For norm-matched synthetic activations, the norm growth starts very high, becomes small, then approaches the natural activation norm growth.
\Cref{fig:errorcorrect} shows synthetic activations moving into norm growth basins.

\subsection{Natural activations inhabit attracting basins}
\label{sec:velocity}

Prior work has shown that language model activations undergo error correction,
in the form of expanding inter-token regions~\citep{shinkle2025activation,pmlr-v282-zavatone-veth26a}.
To systematically study transformer layers' operations on the latent space,
we use the first derivative of our sphere probe \(\dot\eta(\theta)\).
Large values of \(\lVert\dot\eta(\theta)\lVert_2\) mean that a small change in input angle results in a relatively large change in output, indicating expansion within the space;
while small values indicate compression, where nearby activations become closer (relative to surrounding regions) along the tangent direction.

Plotting \(\lVert\dot\eta(\theta)\rVert_2\) in \cref{fig:great_circle_speed}
reveals that activations occur within a velocity basin, with velocity peaks on either side at about \ang{\pm90}.
This is consistent with observations from \citet{shinkle2025activation},
and can be interpreted as a result of a corrective mechanism whereby natural activations are more robust to perturbation 
than norm-matched random activations at \ang{\pm90}.

\begin{figure}
  \centering
  \begin{minipage}[t]{\dimexpr(\textwidth-0.1in)/2\relax}
  \vspace{0pt}
  \includegraphics{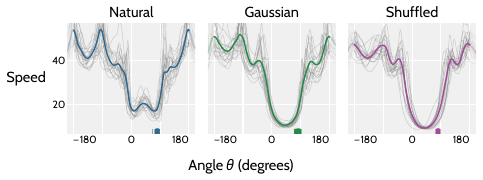}
  \caption{
    \textbf{Speed profiles for natural and synthetic activations} (Llama layers 16--19,
    other models \cref{fig:great_circle_pair_speed_appendix}.)
    Colored curves average over 128 pairs. \ang{0} marks first activation, colored ticks mark paired activation angles, which vary little. 
  }
  \label{fig:great_circle_pair_speed}
  \end{minipage}%
  \hspace{0.1in}%
  \begin{minipage}[t]{\dimexpr(\textwidth-0.1in)/2\relax}
  \vspace{0pt}
  \centering
  \includegraphics{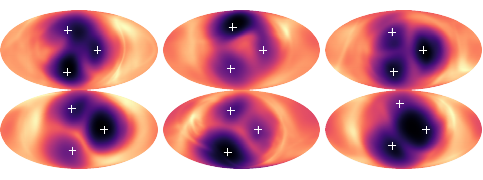}
  \caption{
    \textbf{Sphere probe expansion maps for six randomly sampled sets of 3 activations.}
    Layers 16--21, Mollweide projection to show full sphere.
    White crosses mark activations. Dark regions are low-expansion, light regions high; log scale.
  }
  \label{fig:tripleheatmap}
  \end{minipage}
\end{figure}

Unlike norm growth, velocity at the activation alone does not distinguish natural and synthetic activations (see \cref{sec:velocityauc} for analysis).
However, the picture changes when we look at the velocity profile for spheres through \emph{two} activations, forwarded through multiple layers.
In particular, we see in \cref{fig:great_circle_pair_speed} that natural activations are separated by a ``velocity barrier''o similar to the observation by \citet{heimersheim2024interim}.
Synthetic activations, despite falling within a velocity basin like the natural activations,
do not exhibit a velocity barrier.
Viewing velocity barriers as repelling/stretching regions, this means that the model dynamics actively separate natural activation subregions,
while synthetic activations generally live in a shared basin.
To validate our observations, we sample a fresh set of 128 pairs of natural activations,
and identify the velocity barrier between them by finding the minimum of the local basin for each activation,
and calculate \textsc{auroc} for a classifier based on the height of the barrier between the basins (with 0 for activations within a shared basin).
The results in \cref{tab:pair_barrier_validation} show high separation (0.938--0.996) for all models.

We can visualize the space between up to three random natural activations 
by initializing a 2D sphere probe that passes through the direction of all three,
and using a heatmap to color code the surface by area, i.e., amount of stretch at the output.
\Cref{fig:tripleheatmap} shows several examples from OLMo.
The attracting basins pack together within one hemisphere.

\subsection{Qualitative uniqueness of natural activation space geometry}

\begin{figure}
  \centering
  \includegraphics{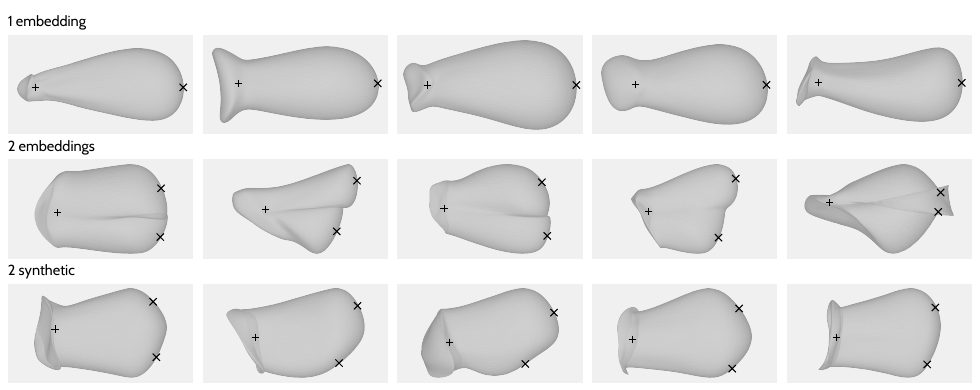}
  \caption{
    \textbf{Projections of 2D spheres after passing through layers 16--20.}
    Spheres are initialized to pass through a single or a pair of 
    activation directions.
    Activation marked with a $\times$, origin marked with a $+$.
    The first and second rows show natural activations; the third row shows shuffled activations.
    The activation space near pairs of natural activations has a distinctive double-lobed profile.
    }
    \label{fig:bubbles}
\end{figure}

The shape of the transformed surface itself reveals a difference between pairs of natural and synthetic activations.
To see the difference, we visualize in \cref{fig:bubbles} the block's output when given a 2D sphere centered at the origin and passing through the activations.
We project this onto a plane defined by the activations and their tangents; see \cref{sec:bubbleimage} for details.
For a sphere passing through a single activation with random tangents, the projection generally presents as a collapsed closed surface at the origin, with a protrusion extending outward from the origin toward the activation.
This is expected, since a random transformation in \(\mathbb{R}^d\) will send most points on the sphere in a direction approximately orthogonal to the projection, collapsing them toward the origin, leaving only the sphere points explicitly preserved by the projection,
i.e., the activation.
The difference between natural and synthetic activations becomes apparent when looking at pairs of activations,
where we see spheres through natural activations exhibit a double-lobed protrusion, with one lobe per activation, 
while synthetic pairs show only a single lobe.

\section{Steering transports activations across basin boundaries}

Language model steering involves perturbing activations along a fixed direction in order to change the behavior of the model~\citep{subramani-etal-2022-extracting,turner2023steering,marks2023geometry,pmlr-v235-park24c}.
Results from \cref{sec:velocity} show that natural activations inhabit separate attracting basins that tile the activation space.
Our hypothesis predicts that model behavior is resilient to interventions like steering because successive layers error-correct the perturbed activations back into natural basins. 
Moreover, this intuition suggests that steering works by moving activations between different basins.
We test this idea by exploring the latent space geometry under two steering tasks: refusal and language steering.

\subsection{Identifying and steering between the refusal and compliance basins}

Previous work in language model alignment has found that refusal is mediated by one or more directions in activation space~\citep{arditi2024refusal, pmlr-v267-wollschlager25a}.
Perturbing activations in the refusal direction can cause the model to refuse benign requests, citing ethical concerns.
Generally, steering is applied to every token position at a specific layer,
in both input and output tokens.
We replicate this method, obtaining a steering vector for \model{Llama 3 8B Instruct} by taking a set of harmful and harmless requests and finding the difference-in-means of the activation for the end-of-user-message token.

\begin{figure}
  \centering
  \includegraphics{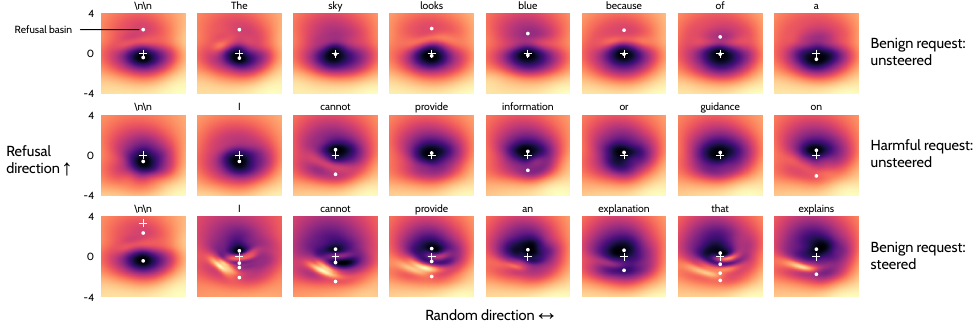}
  \caption{
    \textbf{Activation space expansion around activations from language model responses} to benign (``Why is the sky blue?'')
    and harmful (``How do I poison someone?'') requests. 
    Vertical axis corresponds to the refusal steering direction,
    horizontal axis is chosen randomly.
    Crosses mark activation positions within the space.
    Along the vertical axis, the maps exhibit one or more expansion basins, with local minima marked with points.
    Activations from compliant responses fall within the lower basin (in the compliance direction)
    while activations from refusal responses fall within the upper basin (in the refusal direction).
    Steering the first activation of a response to a benign request into the refusal basin results in a refusal,
    with subsequent tokens falling in the refusal basin.
  }
  \label{fig:refusalsteer}
\end{figure}

We visualize activation space expansion along the refusal steering direction at every token position for a response to a benign request in the first row of \cref{fig:refusalsteer}.
The heatmap shows a basin at each token activation, with many showing a second, shallow basin in the refusal direction.
The second row shows the response to a harmful request, which elicits a refusal from the model.
Mirroring the benign request, many of the heatmaps show a pair of basins---a deep basin where the response activations lie, and a shallow one in the anti-steering direction.
In the third row, we give the model the same benign request,
and steer only the first token position of the response, 
moving the activation into the refusal basin.\footnote{We steer the activation to the \emph{speed} minimum along the steering direction, which does not necessarily coincide exactly with an expansion minimum in \cref{fig:refusalsteer}, which incorporates a random perturbation direction.}
This triggers a different set of expansion maps in the response:
most heatmaps show multiple basins,
and the activations tend to inhabit the upper basins.

Our refusal steering method deviates from that of \citet{arditi2024refusal}.
In particular, refusal steering is typically applied at low strength at every token position in both the request and the response.
We apply steering at a relatively higher strength at a single token position, but the behavioral effect is the same. 
We do so in order to demonstrate the effect of steering between basins, but this also suggests a less-intrusive alternative to steering at all positions.

\subsection{Steering between language feature basins for greater behavioral control}

In order to quantify the role of activation basins in language model inference,
we turn to steering between languages with different scripts, since this allows us to easily determine from a single token which language the model is generating in.\footnote{This is a rough measure, since some languages can be written in multiple scripts, and many languages share scripts, but it suffices for our purposes.}
In particular, we select English (\texttt{en}), Greek (\texttt{el}), and Chinese (\texttt{zh}).
We obtain steering vectors between these languages by finding the activation centroid at layer 16 for a set of documents from each language's \dataset{Wikipedia} corpus, and subtracting the centroids from one another~\citep{sterz-etal-2025-recover}.
We will refer to the vector that steers from the \texttt{en} centroid to the \texttt{zh} centroid as the \texttt{en}--\texttt{zh} vector and so on.
Adding one of these steering vectors, say the \texttt{en}--\texttt{zh}, to layer 16 \texttt{en} activations
will often cause the model to predict the next token as a Chinese character.
This simple setup is remarkably effective: on \num{10000} held out tokens
a steered Llama 3.1 model produced non-Chinese characters only 17\% of the time
and Chinese characters 59\% of the time 
(the remaining generations were  non-language-specific).
Tuning the magnitude of the steering vector to minimize non-Chinese generations further increases the margins to 11\% non-Chinese and 68\% Chinese.

\begin{figure}
  \centering
  \includegraphics{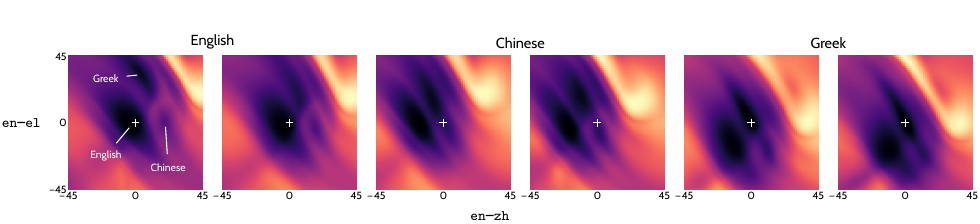}
  \caption{
    \textbf{Activation space expansion along \texttt{en}--\texttt{zh} (horizontal axis) and \texttt{en}--\texttt{el} (vertical axis) tangent directions}
    for 2D sphere probes through activations from English, Chinese, and Greek Wikipedia documents, two each. 
    Colors indicate the relative expansion of the tangent plane to the sphere probe,
    as it passes through Llama layers 16--19, with lighter colors indicating more expansion, log scale.
    Activations from each language inhabit their own low-expansion basin (dark regions).
  }
  \label{fig:language_expansion_maps}
\end{figure}

The latent space geometry along the language feature directions reveals a set of language-specific basins:
using orthogonalized \texttt{en}--\texttt{zh} and \texttt{en}--\texttt{el} steering vectors as the tangent plane for a 2D sphere probe and plotting the resulting expansion
around activations from each language in \cref{fig:language_expansion_maps},
we see a distinct pattern of three basins, one for each language, across all samples.
Token activations from each language fall into their respective basins.
From visual inspection alone,
it is possible to determine which language each activation is from.
Further, the geometric structure of the basins suggests a hierarchical treatment of languages within the model:
English, the model's predominant language during training, appears as the largest basin among the three;
unlike the more equally divided regions seen in \cref{fig:tripleheatmap},
the dividing barriers between the language basins have a $\top$ shape.

\begin{figure}
  \centering
  \includegraphics{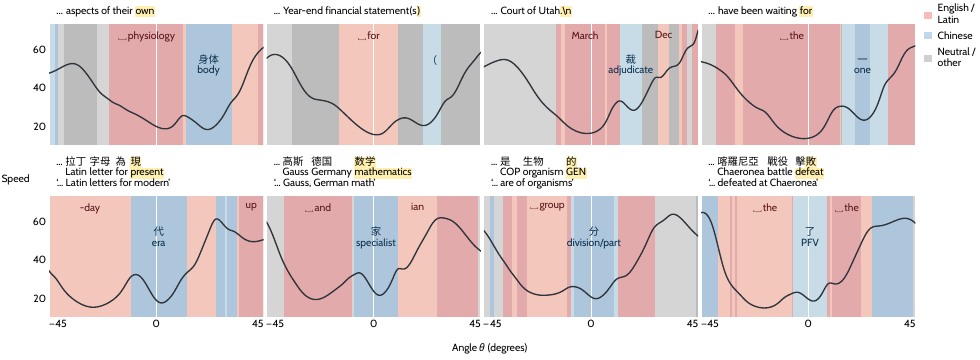}
  \caption{
    \textbf{Speed profiles along the \texttt{en}--\texttt{zh} steering direction} for \texttt{en} (top) and \texttt{zh} (bottom) documents in Llama layers 16--19.
    Regions are color coded by the script type of the model's greedy output, alternating between light and dark to indicate boundaries within a single type.
    Neutral tokens include punctuation and the like.
    The token in the second panel is a fullwidth parenthesis used in Chinese.
  }
  \label{fig:language_steering_speed}
\end{figure}

The language basins also reflect model behavior under steering.
We apply our sphere probe along the \texttt{en}--\texttt{zh} direction to \texttt{en} and \texttt{zh} activations with ablated context,
and record the token type of the model's greedy output under the perturbed activation.
The result in \cref{fig:language_steering_speed} shows a consistent pair-of-basins pattern. 
Activations from the first basin decode to \texttt{en} tokens,
while activations from the second decode to Chinese tokens.
This observation suggests an improved method for calibrating steering strength.
We hypothesize that steering works by moving activations from one basin to another.
Under this hypothesis, steering failures can occur when the steering vector over- or under-shoots the basin.
We therefore predict that modulating the steering strength to optimally place tokens at the velocity minimum of the neighboring basin should improve steering effectiveness.
We test this prediction by comparing our adaptive method to fixed-strength steering on the set of held out activations in \cref{tab:adaptive_steer}.

\begin{table}
  \centering
  \begingroup
\small
\sisetup{detect-weight=true}
\setlength{\tabcolsep}{3pt}
\begin{tabular}{lS[table-format=2.2]S[table-format=2.2]S[table-format=2.2]S[table-format=2.2]|S[table-format=2.2]S[table-format=2.2]S[table-format=2.2]S[table-format=2.2]}
\toprule
{} & \multicolumn{4}{c}{non-\texttt{zh} $\downarrow$} & \multicolumn{4}{c}{\texttt{zh} $\uparrow$} \\
{} & \multicolumn{2}{c}{Mass (\%)} & \multicolumn{2}{c}{Greedy (\%)} & \multicolumn{2}{c}{Mass (\%)} & \multicolumn{2}{c}{Greedy (\%)} \\
{Method} & {\textsc{id}} & {\textsc{ood}} & {\textsc{id}} & {\textsc{ood}} & {\textsc{id}} & {\textsc{ood}} & {\textsc{id}} & {\textsc{ood}} \\
\midrule
Untuned & 20.10 & 21.48 & 17.16 & 18.74 & 61.63 & 63.82 & 59.23 & 58.61 \\
Tuned & 13.82 & 10.43 & 11.18 & 8.18 & 69.49 & 77.62 & 68.07 & 74.81 \\
Adaptive & \bfseries 12.81 & \bfseries 9.17 & \bfseries 10.55 & \bfseries 6.61 & \bfseries 70.69 & \bfseries 79.49 & \bfseries 69.10 & \bfseries 77.94 \\
Oracle & 9.39 & 7.66 & 4.43 & 3.92 & 73.19 & 80.32 & 79.13 & 82.36 \\
\bottomrule
\end{tabular}
\endgroup

  \caption{
    Steering effectiveness of various strength-selection methods,
    as measured by mean probability mass on---and greedy decoding rate of---\texttt{zh} and non-\texttt{zh} tokens for in- and out-of-distribution test sets.
    The untuned \texttt{en}--\texttt{zh} centroid difference serves as a baseline,
    improved upon by tuning the fixed steering strength on a development set.
    Adaptive steering, our method, selects steering strength on a per-instance basis
    to land the perturbed activation within a velocity basin.
    Oracle steering, which selects a per-instance steering strength to minimize non-\texttt{zh} tokens,
    upper bounds performance.
  }
  \label{tab:adaptive_steer}
\end{table}

We measure steering effectiveness in two ways: probability of greedily decoding a Chinese or non-Chinese token (excluding language-ambiguous tokens),
and the total probability mass on Chinese and non-Chinese tokens.
For fixed-strength steering, we tune the strength on a set of \num{1000} activations from \texttt{en} \dataset{Wikipedia} to minimize either greedy decodes of non-Chinese tokens or the total probability mass on non-Chinese tokens.
As an upper bound for performance, 
an oracle steering method selects the optimal steering strength independently for each test instance.
The oracle does not achieve perfect steering because some activations,
mainly partial word completions, do not produce non-\texttt{en} tokens regardless of steering strength.
For our method, we find the local velocity basin containing the unperturbed activation, then set the strength to land the perturbed activation exactly at the deepest velocity minimum in the steered direction, up to a maximum steering strength.
If no second minimum occurs, we check if there is a second minimum in the opposite direction (which often happens when the output is already in Chinese),
in which case we apply no steering. If there is no minimum in either direction, we fall back to the tuned fixed steering strength (this occurs less than 0.5\% of the time in all of our evaluation sets).
We test on a held-out set of \num{10000} tokens from \texttt{en} \dataset{Wikipedia},
and another \num{10000} from \dataset{FineWeb} (excluding Wikipedia) to test out-of-distribution (\textsc{ood}) generalization.

Across the board, adaptive steering outperforms fixed steering methods, closing the gap in many cases to within a few percentage points of the upper bound.
All differences are significant under 95\% paired CI.
Notably, the gap between tuned fixed-strength steering and adaptive steering is larger on out of distribution tokens than in-distribution,
suggesting that adaptively steering into feature basins is a more generalizable strategy.
Whereas earlier experiments were largely observational, our steering experiment demonstrates that feature basins play an active role in neural computation, and shows that our insights can lead to practical gains in language model control.

\section{Conclusion}

Our investigation began with an intuition about how geometric constraints in activation space---like biomechanical constraints in living organisms---could contribute to robust and desirable behavior in language models.
Based on this premise, we employed probes to investigate the activation space geometry.
Our observations allowed us to characterize the activation space geometry as a set of attracting basins that exhibit error correction, forming privileged regions where natural activations reside.
Armed with this understanding, we were able to gain insights into model steering,
observing feature-specific basins in refusal and multilingual steering directions,
whose structure suggests hierarchical representation within the model.
Hypothesizing that ideal steering moves activations between feature basins,
we develop a basin-aware, parameter-free steering method and show that it improves over fixed-strength steering in the English-Chinese steering setting.
This result establishes the role of these feature basins in model behavior,
and recommends analysis of activation space geometry through layer dynamics as a fruitful and practically useful lens for interpretability.

This work is largely exploratory, 
and as a consequence, we leave many open questions and ideas for future work.
For instance, we anticipate that the geometric picture becomes more complex when considering very similar activations, e.g., activations for the same token in slightly different contexts, which may occupy a low-dimensional manifold.
We also anticipate that feature basins could play a role in circuits and more complex computation in models.
Lastly, we are excited by the prospect of exploring the activation space without relying on data. Our work focused on feature directions and activations obtained from data, but new methods~\citep[e.g.,][]{mack2024melbo} could automatically find new basins or other consequential geometric features within the space to predict how the model will behave and how it computes.

\section{Acknowledgements}

Our thanks to Stefan Heimersheim for their draft feedback and discussions.
We also thank the team at Goodfire for their feedback and encouragement during the research process.
The University of Southern California Natural Language Processing group, including \textsc{dill} and \textsc{ink} labs, also provided valuable comments on the abstract, introduction section, and \cref{fig:coast}.
Matthew Finlayson's PhD work is supported by a National Science Foundation Graduate Research Fellowship (\textsc{nsf grfp}).
Xiang Ren’s research is supported in part by the Office of the Director of National Intelligence (\textsc{odni}), Intelligence Advanced Research Projects Activity (\textsc{iarpa}), via the \textsc{hiatus} Program contract \#2022-22072200006, the Defense Advanced Research Projects Agency with award HR00112220046, and \textsc{nsf iis} 2048211.

\SilicoAcknowledgement

\clearpage
\bibliographystyle{plainnat}
\bibliography{main}

\clearpage
\appendix

\section{Additional experimental results}
\label{sec:norm_growth_context}

\begin{figure}[H]
  \centering
  \includegraphics{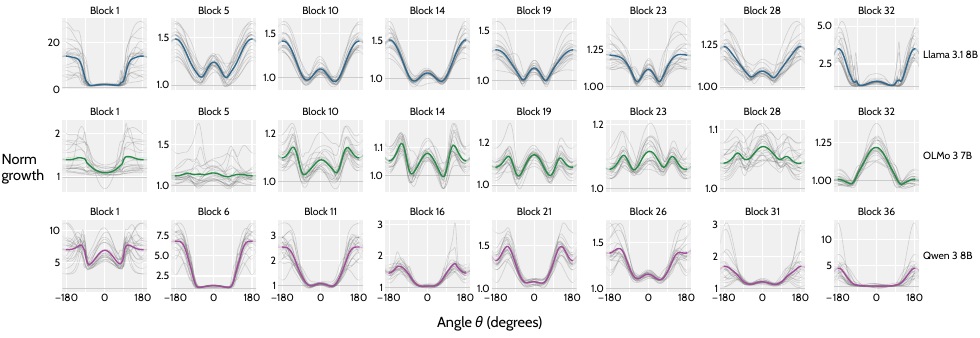}
  \caption{
    Norm growth profiles with non-ablated context.
  }
  \label{fig:great_circle_norm_growth_context}
\end{figure}

\begin{table}[H]
  \centering
  \begingroup
\scriptsize
\setlength{\tabcolsep}{2pt}
\begin{tabular}{llS[table-format=1.3, round-mode=places, round-precision=3]S[table-format=1.3, round-mode=places, round-precision=3]S[table-format=1.3, round-mode=places, round-precision=3]S[table-format=1.3, round-mode=places, round-precision=3]S[table-format=1.3, round-mode=places, round-precision=3]S[table-format=1.3, round-mode=places, round-precision=3]S[table-format=1.3, round-mode=places, round-precision=3]S[table-format=1.3, round-mode=places, round-precision=3]S[table-format=1.3, round-mode=places, round-precision=3]S[table-format=1.3, round-mode=places, round-precision=3]S[table-format=1.3, round-mode=places, round-precision=3]S[table-format=1.3, round-mode=places, round-precision=3]S[table-format=1.3, round-mode=places, round-precision=3]S[table-format=1.3, round-mode=places, round-precision=3]S[table-format=1.3, round-mode=places, round-precision=3]}
\toprule
{} & {Layer} & {1} & {4} & {6} & {8} & {11} & {14} & {16} & {18} & {21} & {24} & {26} & {28} & {31} & {34} & {36} \\
{Class} & {Model} & {} & {} & {} & {} & {} & {} & {} & {} & {} & {} & {} & {} & {} & {} & {} \\
\midrule
Gaussian & Llama 3.1 8B & 0.851 & 0.963 & 0.915 & 0.897 & 0.789 & 0.818 & 0.963 & 0.983 & 0.981 & 0.899 & 0.869 & 0.759 & 0.555 & {\textemdash} & {\textemdash} \\
 & OLMo 3 7B & 0.618 & 0.808 & 0.887 & 0.890 & 0.893 & 0.933 & 0.947 & 0.954 & 0.997 & 0.973 & 0.935 & 0.928 & 0.742 & {\textemdash} & {\textemdash} \\
 & Qwen 3 8B & 0.765 & 0.621 & 0.824 & 0.865 & 0.923 & 0.833 & 0.690 & 0.740 & 0.830 & 0.923 & 0.965 & 0.955 & 0.836 & 0.669 & 0.478 \\
Shuffled & Llama 3.1 8B & 0.927 & 0.999 & 0.991 & 0.956 & 0.980 & 0.999 & 1.000 & 0.998 & 1.000 & 0.980 & 0.950 & 0.928 & 0.835 & {\textemdash} & {\textemdash} \\
 & OLMo 3 7B & 0.525 & 0.887 & 0.954 & 0.913 & 0.927 & 0.956 & 0.974 & 0.986 & 0.990 & 0.975 & 0.903 & 0.902 & 0.866 & {\textemdash} & {\textemdash} \\
 & Qwen 3 8B & 0.997 & 0.906 & 0.969 & 0.908 & 0.943 & 0.953 & 0.871 & 0.957 & 0.995 & 0.998 & 0.995 & 0.981 & 0.944 & 0.786 & 0.503 \\
\bottomrule
\end{tabular}
\endgroup

  \caption{
    For selected layers, we measure \textsc{roc-auc} of a simple, norm growth threshold classifier for whether an activation is natural or synthetic.
    Most layers of most models score high, meaning that the \num{128} natural and synthetic activations in our test set are almost perfectly separable.
  }
  \label{tab:fstorderclassifier}
\end{table}

\begin{figure}[H]
  \centering
  \includegraphics{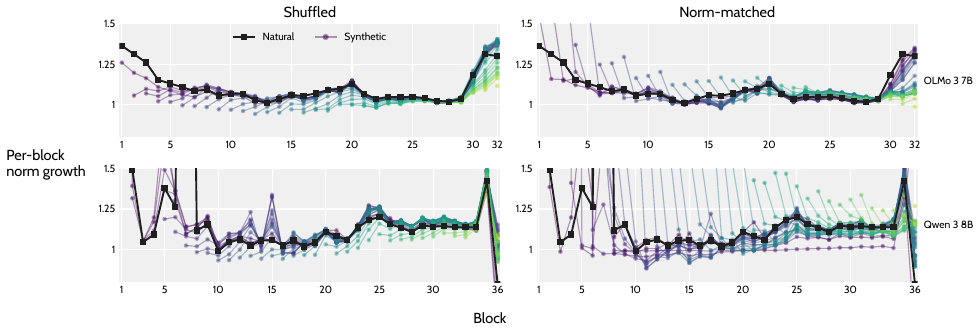}
  \caption{Norm growth trajectories for synthetic activations in OLMo and Qwen.}
  \label{fig:traject}
\end{figure}

\begin{figure}[H]
  \centering
  \includegraphics{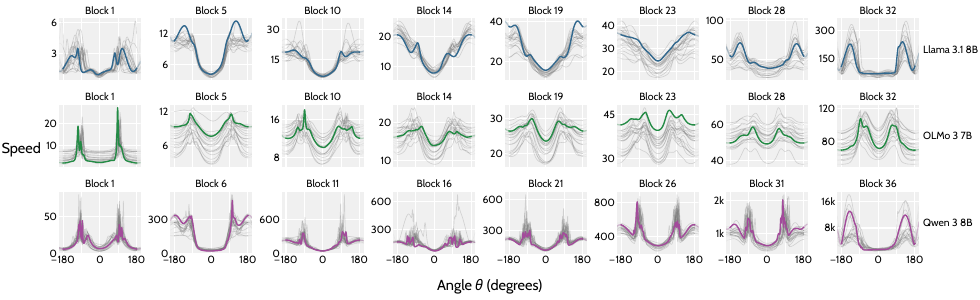}
  \caption{
    \textbf{Per-layer speed around natural activations.} Vertical axis gives speed, i.e., \(\lVert\dot\eta(\theta)\rVert_2\).
  }
  \label{fig:great_circle_speed}
\end{figure}

\label{sec:velocityauc}

\begin{figure}[H]
  \centering
  \includegraphics{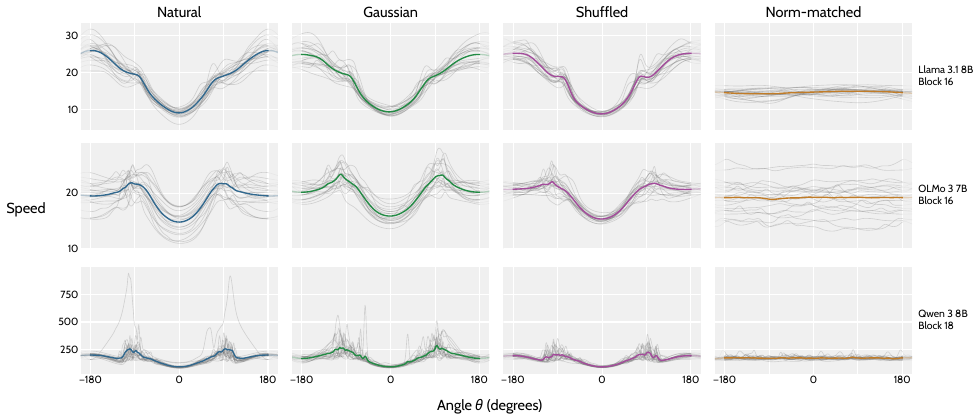}
  \caption{
    Comparing natural to synthetic activations' speed profiles at for various models.
  }
  \label{fig:great_circle_synthetic_speed}
\end{figure}

\Cref{fig:great_circle_synthetic_speed} shows the velocity profiles for single layer transformer blocks. There is no significant difference between natural and synthetic activations in terms of velocity profile for just a single layer.

\label{sec:pair_barriers_additional}

\begin{table}[H]
  \centering
  \begin{tabular}{lS[table-format=1.3, round-mode=places, round-precision=3]S[table-format=1.3, round-mode=places, round-precision=3]S[table-format=1.3, round-mode=places, round-precision=3]}
\toprule
Synthetic class & {Llama 3.1 8B} & {OLMo 3 7B} & {Qwen 3 8B} \\
\midrule
Gaussian & 0.963 & 0.984 & 0.938 \\
Shuffled & 0.980 & 0.996 & 0.986 \\
\bottomrule
\end{tabular}

  \caption{
    Velocity barrier--based classification of synthetic and natural activations.
    All models and synthetic activation types show high separation.
  }
  \label{tab:pair_barrier_validation}
\end{table}

\begin{figure}[H]
  \centering
  \includegraphics{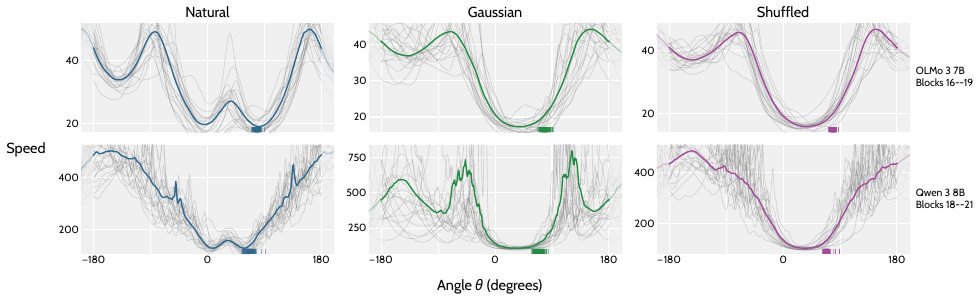}
  \caption{Barriers between natural activations in OLMo 3 7B and Qwen 3 8B,
    complementing the Llama results in \cref{fig:great_circle_pair_speed}.
    Columns compare natural, Gaussian, and coordinate-shuffled pairs.
    Colored curves show means over 128 pairs; gray curves show individual examples.}
  \label{fig:great_circle_pair_speed_appendix}
\end{figure}

\section{Details for 2D sphere images}
\label{sec:bubbleimage}

For single activations~\(\vx\), we choose two random tangents at \(\vx\) to define the sphere.
We project the surface onto the plane spanned by \(f(\vx)\) and \(\dot\eta(0)\),
where \(\eta(\theta)\) parameterizes the 1D sphere along one of the tangents.
For pairs of activations \(\vx,\vx'\),
we scale \(\vx'\) to have the same norm as \(\vx\),
then define the sphere with one tangent~\(\vu\) at \(\vx\)
toward the second activation~\(\vx'\), i.e., \(\vu=(\vx-\vx')\perp\vx\),
and choose a random second tangent.
For the projection parameterize a 1D sphere probe with \(\eta(0)=f(\vx)\) and \(\eta(\phi)=f(\vx')\),
and project the transformed 2D sphere onto the plane spanned by \(f(\vx)+f(\vx')\) and \(\dot\eta(0)+\dot\eta(\phi)\).

\end{document}